\documentclass[conference]{./IEEEtran}
\IEEEoverridecommandlockouts
\usepackage{cite}
\usepackage[cmex10]{amsmath}
\usepackage{algorithmic}
\usepackage{graphicx}
\usepackage{textcomp}
\usepackage{xcolor}

\usepackage{multirow}
\usepackage{subfigure}
\usepackage{array}
\usepackage{balance}

\begin{document}

\title{Cross-Domain Labeled LDA for Cross-Domain Text Classification}

\author{\IEEEauthorblockN{Baoyu Jing\IEEEauthorrefmark{1}$^1$\thanks{$^1$ Equal contribution.}, 
Chenwei Lu\IEEEauthorrefmark{2}$^1$, 
Deqing Wang\IEEEauthorrefmark{2}$^2$\thanks{$^2$ Corresponding author.}, 
Fuzhen Zhuang\IEEEauthorrefmark{3}\IEEEauthorrefmark{4}, and
Cheng Niu\IEEEauthorrefmark{5}}
\IEEEauthorblockA{\IEEEauthorrefmark{1}School of Computer Science, Carnegie Mellon University, Pittsburgh, PA, USA}
\IEEEauthorblockA{\IEEEauthorrefmark{2}School of Computer Science and Engineering, Beihang University, Beijing, China}
\IEEEauthorblockA{\IEEEauthorrefmark{3}Key Lab of Intelligent Information Processing of Chinese Academy of Sciences (CAS)\\ Institute of Computing Technology, CAS, Beijing, China}
\IEEEauthorblockA{\IEEEauthorrefmark{4}University of Chinese Academy of Science, Beijing, China}
\IEEEauthorblockA{\IEEEauthorrefmark{5}Pattern Recognition Center, WeChat Search Application Department, Tencent, China}
Email: byjing@cs.cmu.edu, \{luchenwei, dqwang\}@buaa.edu.cn, zhuangfuzhen@ict.ac.cn, niucheng@tencent.com}

\maketitle

\begin{abstract}
Cross-domain text classification aims at building a classifier for a target domain which leverages data from both source and target domain. One promising idea is to minimize the feature distribution differences of the two domains. Most existing studies explicitly minimize such differences by an \emph{exact alignment} mechanism (aligning features by one-to-one feature alignment, projection matrix etc.). Such \emph{exact alignment}, however, will restrict models' learning ability and will further impair models' performance on classification tasks when the semantic distributions of different domains are very different. To address this problem, we propose a novel \emph{group alignment} which aligns the semantics at group level. In addition, to help the model learn better semantic groups and semantics within these groups, we also propose a partial supervision for model's learning in source domain. To this end, we embed the \emph{group alignment} and a partial supervision into a cross-domain topic model, and propose a Cross-Domain Labeled LDA (CDL-LDA). On the standard 20Newsgroup and Reuters dataset, extensive quantitative (classification, perplexity etc.) and qualitative (topic detection) experiments are conducted to show the effectiveness of the proposed \emph{group alignment} and partial supervision.
\end{abstract}

\begin{IEEEkeywords}
Cross Domain Text Classification, Topic Modeling, Group Alignment, Partial Supervision
\end{IEEEkeywords}


\section{Introduction}\label{Intro}
Cross-domain text classification considers the setting that data distributions in source domain and target domain are different but related. 
In such a scenario, the performance of traditional classification algorithms, which are built on the assumption that the source and target datasets are drawn from the same distribution~\cite{pan2010survey},  will be deteriorated~\cite{dai2007co,li2012topic,bao2013partially}. Therefore, many cross-domain learning methods are proposed, such as instance-based methods~\cite{jiang2007instance,mansour2009domain,huang2007correcting}, co-training \cite{wan2009co,chen2011co}, kernel methods~\cite{duan2012domain,wang2017fredholm} and representation learning~\cite{ben2007analysis,gupta2010nonnegative,wang2014cross,glorot2011domain,zhai2004cross,xue2008topic,zhuang2010collaborative,paul2009cross,bao2013partially,li2012topic,shen2017adversarial,ganin2016domain,wei2016deep,jiang2016l2,ICML2012Chen_416}. 

Most existing representation learning models try to align features of the two domains through an \emph{exact alignment}, which aligns features of the two domains by a shared feature space~\cite{gupta2010nonnegative, wang2014cross,shen2017adversarial,ganin2016domain,wei2016deep,jiang2016l2,ICML2012Chen_416}, one-to-one topic alignment~\cite{xue2008topic,paul2009cross,bao2013partially}, or a projection matrix~\cite{li2012topic}. Generally, this \emph{exact alignment} mechanism assumes that the semantics of the target domain can be directly decomposed by the semantics of the target domain. However, such assumption is not always promised to be true in the real-world data because the numbers and contents of semantics in two domains are always very different. For example, the source domain might be comprised of two topics (\emph{graphics} and \emph{hockey}), and the target domain might consist of three topics 
(\emph{software}, \emph{commands}, and \emph{baseball}). 

The topics in the above example can be clustered into two groups: \emph{computers} (\emph{graphics}, \emph{software}, \emph{commands}) and \emph{recreation} (\emph{hockey}, \emph{baseball}). It is much more intuitive to align topics through topic groups (\emph{computers} and \emph{recreation}) rather than directly align them at topic level. Motivated by this intuition, we propose a novel \emph{group alignment} mechanism. Topic groups can be defined in many ways: cluster of topics, document labels etc. For simplicity, we define groups through document labels. The \emph{group alignment} has two major advantages: 1) Topic groups are guaranteed to exist in both source and target domains if they are predefined by document labels, thus aligning topics by such groups are always feasible. 2) The numbers of topics within different topic groups are allowed to be different, and thus model's representation flexibility for different domains might be improved. 


Additionally, partial supervision for the learning in source domain has been proven to help the learning in target domain~\cite{bao2013partially}. Therefore, to help the model learn better topic groups and topics within each group, we propose a partial supervision for topic learning in source domain. 

To this end, we propose a Cross-Domain Labeled LDA (CDL-LDA) model for cross-domain text classification which is equipped with a novel \emph{group alignment} and a partial supervision. The experiment results on the standard 20Newsgroup and Reuters datasets show that CDL-LDA can achieve higher classification accuracies and lower perplexities than the state-of-the-art models. Besides, CDL-LDA can also detect meaningful topic groups and topics. Additionally, parameter analysis is also conducted to show further characteristics of the model.

\section{Related Work}\label{related work}
\subsection{Cross-Domain Learning}
Cross-domain classification or transductive classification~\cite{pan2010survey} has attracted much attention in recent years. Generally, there are four types of methods: instance re-weighting methods~\cite{jiang2007instance,mansour2009domain,huang2007correcting}, co-training methods~\cite{wan2009co,chen2011co}, kernel methods~\cite{duan2012domain,wang2017fredholm}, and feature representation based methods~\cite{ben2007analysis,blitzer2006domain,prettenhofer2010cross,pan2010cross,gupta2010nonnegative,wang2014cross,glorot2011domain,zhai2004cross,xue2008topic,zhuang2010collaborative,paul2009cross,bao2013partially,shen2017adversarial,ganin2016domain,wei2016deep,jiang2016l2,ICML2012Chen_416}. A comprehensive survey on transfer learning and transductive classification can be found in reference~\cite{pan2010survey}. In this paper, we mainly focus on methods based on high-level semantic features~\cite{gupta2010nonnegative,wang2014cross,glorot2011domain,zhai2004cross,paul2009cross,bao2013partially,shen2017adversarial,ganin2016domain,wei2016deep,jiang2016l2,ICML2012Chen_416}. 

Most popular methods for extracting high-level semantic features include matrix factorization, distance metric learning, deep learning methods, and topic modeling methods. For example, Gupta et al~\cite{gupta2010nonnegative} propose a shared nonnegative matrix factorization (JSNMF) to jointly extract domain-independent and domain-dependent bases. Cross-Domain Metric Learning (CDML)~\cite{wang2014cross} transfers knowledge by finding a shared Mahalanobis distance across domains.

As for deep learning methods, one set of models are based on auto-encoders. Glorot et al~\cite{glorot2011domain} employ a Stacked De-noising Auto-encoder (SDA) to learn the high-level features across different domains in an unsupervised fashion. Chen et al~\cite{ICML2012Chen_416} modify SDA by using linear denoisers and propose marginalized SDA (mSDA). Jiang et al~\cite{jiang2016l2} and Wei et al \cite{wei2016deep} modify mSDA by using $\ell_{2,1}$ norm for objective function and introducing maximum mean discrepancy into mSDA. Zhou et al~\cite{zhou2016bi} propose a Bi-Transferring Deep Neural Networks (BTDNNs) in which different domains have different decoders. Another set of models are based on domain-adversarial neural networks. Ganin et al~\cite{ganin2016domain} propose a Domain-Adversarial Neural Network (DANN) which builds a label predictor and a domain classifier on a feedforward neural network. Recently, Shen et al~\cite{shen2017adversarial} replace domain discriminator of DANN by Wasserstein distance~\cite{ruschendorf1985wasserstein} and propose Adversarial Representation for Domain Adaptation (ARDA). Generally, all these methods project different domains into different regions of a shared feature space, and directly minimize the distance between these regions. However, when the semantics of two domains are very different, such direct minimization might hurt models' representation flexibility.

\subsection{Cross-Domain Topic Models} \label{cc topic models}

One idea of cross-domain topic modeling methods is to divide topics into common topics and specific topics to capture shared semantics and domain specific semantics respectively~\cite{zhai2004cross, paul2009cross, li2012topic, bao2013partially, zuo2015complementary}. To perform text classification, one can choose to use either common topic features or both common and specific topic features. However, common topic features alone might not provide enough information for classification, thus different topic alignment methods are proposed. For example, Cross-collection mixture model (CCMix)~\cite{zhai2004cross} and Cross-Collection Latent Dirichlet Allocation (CCLDA)~\cite{paul2009cross} extends Probabilistic Latent Semantic Analysis (PLSA)~\cite{hofmann1999probabilistic} and Latent Dirichlet Allocation (LDA)~\cite{blei2003latent} into cross-collection topic models respectively, and they align specific topics by an one-to-one alignment. Partially supervised CCLDA (PSCCLDA)~\cite{bao2013partially} extends CCLDA into a partially supervised model by incorporating a logistic regression in the source domain at the training process. Topic Correlation Analysis (TCA)~\cite{li2012topic} assumes that the specific topics of the target domain can be decomposed by the specific topics of the source domain, and thus aligns specific topics across different domains by a projection matrix. Recently, Zuo et al~\cite{zuo2015complementary} propose a fine-grained Cross-collection Auto-labeled MaxEnt-LDA (CAMEL), which further divides topics into opinions and aspects while it still employs an one-to-one topic alignment to align specific topics.

\section{Proposed Model}
\subsection{Problem Definition}
Given a target domain dataset $\mathcal{D}^t=\{\mathbf{x}^t_1,\cdots,\mathbf{x}^t_{N_t}\}$ containing $N_t$ unlabeled documents, and a source domain dataset $\mathcal{D}^s=\{(\mathbf{x}^s_1,y^s_1), \cdots, (\mathbf{x}^s_{N_s},y^s_{N_s})\}$ containing $N_s$ labeled documents. Here $\mathbf{x}^s_i$ and $y^s_i$ are the feature vector and the label of the $i$-th example in the source domain $\mathcal{D}^s$ respectively, and $\mathbf{x}^t_i$ is the feature vector of the $i$-th example in the target domain $\mathcal{D}^t$. Let $\mathcal{Y}$ be the predefined set of document labels, then our task is to train a classifier for the target domain: $f^t:\mathcal{D}^t\rightarrow\mathcal{Y}$.

\subsection{Overview of Cross-Domain Labeled LDA}
Cross-domain Labeled LDA (CDL-LDA) is a cross-domain topic model which divides topics into common topics and specific topics to model shared semantics across domains and domain dependent semantics respectively. To perform text classification, it is necessary to align specific topics across domains. However, different from \emph{exact alignment} (one-to-one topic alignment, projection matrix etc.) adopted by previous studies~\cite{zhai2004cross, paul2009cross, li2012topic, bao2013partially, zuo2015complementary}, which directly performs topic alignment at topic level, we propose a novel \emph{group alignment} which performs topic alignment at topic group level. To be more specific, common topics and specific topics are firstly divided into different groups. Then within each domain, common topics and specific topics of the same group are aligned. Finally, across domains, specific topics of the same topic group are aligned through common topics in this topic group. We will elaborate \emph{group alignment} in section~\ref{group alignment}.

In addition to \emph{group alignment}, we also propose a partial supervision to explicitly incorporate ground truth topic group information in the source domain, which can help the model learn better topic features for classification. To do so, instead of sampling topic group labels for words in source domain, the model directly assign ground truth topic group labels of the words at training time. We will discuss more about the partial supervision in section~\ref{supervision}.




Additionally, there are several basic assumptions in CDL-LDA. Firstly, each document has a multinomial distribution over topic group labels $\pi$ which is drawn from symmetric Dirichlet prior $Dir(\eta)$. Secondly, given topic group label $l$ sampled from $\pi$, topic type (common/specif) switcher $r$ is modeled by a Bernoulli distribution $\sigma_l$ which is drawn from symmetric beta prior $Beta(\gamma)$. Then, given label $l$ and topic type $r$, topic $z$ is modeled by a multinomial distribution $\theta^C$ ($r=0$) or $\theta^S$ ($r=1$), which is drawn from $Dir(\alpha)$. Finally, each topic is modeled by a multinomial distribution $\phi$, which has the size of $V$. $\phi$ can be further divided into common topics $\phi^C$ and specific topics $\phi^S$. For simplicity, we assume $\phi^C$ and $\phi^S$ are drawn from same symmetric Dirichlet prior $Dir(\beta)$. 

\subsection{Generative Process of Cross-Domain Labeled LDA}

\begin{enumerate}
  \item[1.] For each label and common topic pair $(l,c)$:
  \begin{enumerate}
    \item[a)] Choose $\phi^C_{l,c}\sim Dir(\beta)$
  \end{enumerate}

  \item[2.] For each collection $m$ (source or target):
  \begin{enumerate}
    \item[a)] For each label and specific topic pair $(l,s)$:\\
    Choose $\phi^S_{m,l,s}\sim Dir(\beta)$
  \end{enumerate}

  \item[3.] For each document $d$:
  \begin{enumerate}
    \item[a)] Choose a domain indicator $m$ \\(not shown in the graph)
    \item[b)] Choose label distribution $\pi_d\sim Dir(\eta)$
    \item[c)] Choose topic distributions: 
    \item[]   common topic distribution $\theta^C_{d,l}\sim Dir(\alpha)$
    \item[]   specific topic distribution $\theta^S_{d,l}\sim Dir(\alpha)$
    \item[d)] Choose topic type distribution\\ $\sigma_{d,l}\sim Beta(\gamma)$ for each label $l$

    \item[e)] For each word $w$ in $d$:
    \begin{enumerate}
      \item[i] Choose label $l\sim Mult(\pi_d)$
      \item[ii] Choose topic type switcher\\ $r_{d,w}\sim Bern(\sigma_{d,l})$
      \item[iii] If $r_{d,w}=0$, choose $z_{d,w}\sim Mult(\theta^C_{d,l})$
      \item[] If $r_{d,w}=1$, choose $z_{d,w}\sim Mult(\theta^S_{d,l})$
    \end{enumerate}
  \end{enumerate}
\end{enumerate}

\begin{figure}[b!]
  \centering
  \footnotesize
  \includegraphics[width=0.4\textwidth]{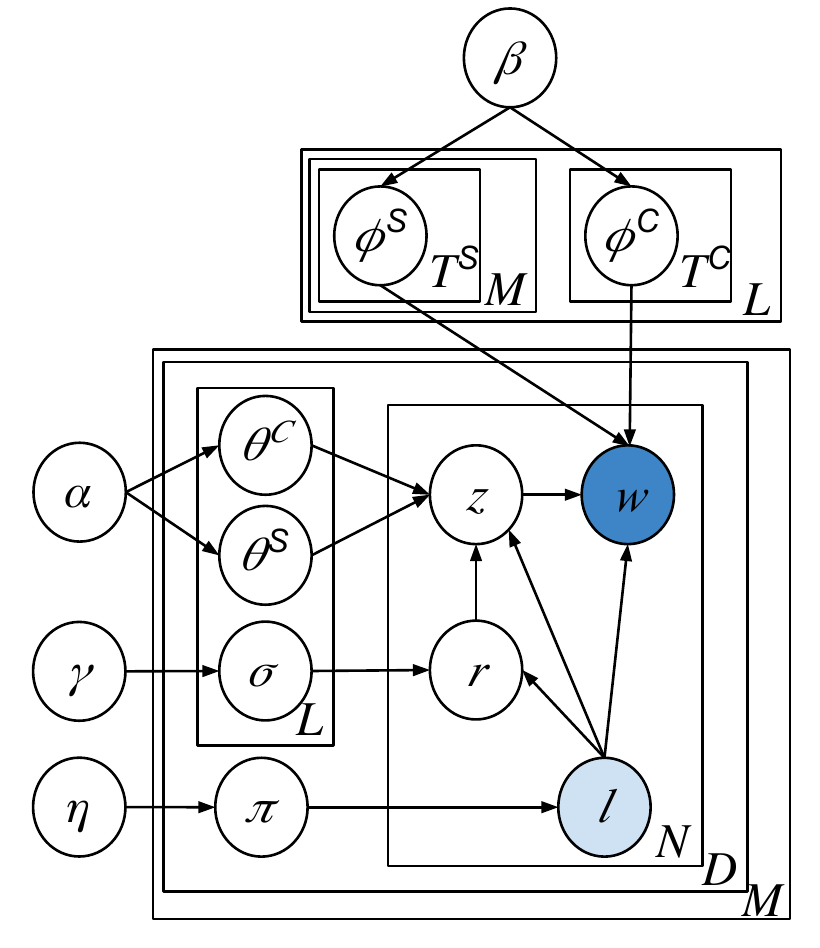}
  \caption{Graphical representation of CDL-LDA}
  \label{Model}
\end{figure}

\begin{table}[!t]
\centering
\caption{Math notations for CDL-LDA and inference.}
\begin{tabular}{ll}
\hline
Notation& Description\\
\hline
$M$ & Number of domains\\
$D$ & Number of documents\\
$N_d$ & Number of words for a document\\
$L$ & Number of labels\\
$T^C$ & Number of common topics\\
$T^S$ & Number of specific topics\\
$m$ & Domain indicator\\
$c$ & Common topic index\\
$s$ & Specific topic index\\
$w$ & An observed word\\
$z$ & Topic assigned for a word\\
$r$ & Topic type (common/specific) switcher\\
$l$ & Label (group) assigned for a document $d$\\
$\phi^C$ & Common topic distribution\\
$\phi^S$ & Specific topic distribution\\
$\theta^C$ & Common topic mixture of $d$\\
$\theta^S$ & Specific topic mixture for $d$\\
$\sigma$ & Distribution for topic type (common/specific) switcher $r$\\
$\pi$  & Topic group distribution\\
$\alpha$ & Dirichlet prior for $\theta^C$/$\theta^S$\\
$\beta$ & Dirichlet prior for $\phi^C$/$\phi^S$\\
$\gamma$ & Beta priors for $\sigma$\\
$\eta$ & Dirichlet prior for topic group distribution $\pi$\\
$Bern(\cdot)$& Bernoulli distribution with parameter($\cdot$)\\
$Beta(\cdot)$& Beta distribution with parameter($\cdot$)\\
$Multi(\cdot)$& Multinomial distribution with parameter($\cdot$)\\
$Dir(\cdot)$& Dirichlet distribution with parameter($\cdot$)\\
\hline
$V$& Vocabulary size\\
$r$& Topic type: 0-common topic; 1-specific topic\\
$N_d$& Number of words in document $d$\\
$N_{l,d}$& Number of words in $d$ assigned with group label $l$\\
$N_{r,l,d}$& Number of words in $d$ assigned with $l$ and topic type $r$\\
$N_{z,r,l,d}$& Number of words in $d$ assigned with $l$, $r$ and topic $z$\\
$N_{z,r,l}$& Number of words assigned with $l$, $r$ and $z$ in the corpus\\
$N_{w_t,z,r,l}$& Number of times word $w_t$ is assigned with $l$, $r$ and $z$\\
\hline
\end{tabular}
\label{tab:notations}
\vspace{-0.3cm}
\end{table}

\subsection{Group Alignment}\label{group alignment}
Most of previous works adopt \emph{exact alignment} to align specific topics across different domains since they assume that the specific topics in the target domain can be decomposed by one \cite{bao2013partially,zhai2004cross,paul2009cross} or several specific topics \cite{li2012topic} in the source domain. However, such assumption might be too strict, since the semantic structures of different domains are always different (e.g. the numbers and contents of topics may vary for different domains), and a topic in the target domain may not have strong correlated topics in the source domain. However, similar or even same topic groups usually exist in both of the source and the target domains, especially when the two domains share the same set of document labels (an example has been provided in section~\ref{Intro}). Motivated by this observation, we propose a novel \emph{group alignment} to relax the assumption of \emph{exact alignment}.

Topic groups can be defined in many ways (e.g. semantic similarity, keywords similarity). In this paper, for simplicity, we define topic groups through document labels. In cross-domain binary classification task, each document must be assigned with a \emph{positive} or \emph{negative} label. Therefore, it is intuitive to assume that the topics can be clustered into \emph{positive} and \emph{negative} groups. Similarly, for cross-domain multi-class classification tasks, topics can be clustered into $L$ groups, where $L$ is the number of document classes. 

There are two advantages of such \emph{group alignment}: 1) the groups are guaranteed to exist in both of the source and the target domains, thus aligning topics by groups are always feasible. 2) the numbers of topics within different groups are allowed to be different, and thus the model will have more flexibility for modeling topics in different domains.

The \emph{group alignment} is embedded in the generative process of CDL-LDA by assuming that topic $z$ and common/specific topic switcher $r$ are generated after label $l$ has been chosen. To be more concrete, within each domain, the model explicitly clusters common and specific topics into different topic groups by choosing topic type switcher $r$ and topic $z$ after label $l$ is determined. Note that common topics of different domains are naturally aligned since common topics are shared by all domains. As a result, across different domains, specific topics belonging to the same topic group are aligned through common topics within this topic group.

\subsection{Supervision}\label{supervision}
Unsupervised models usually ignore the valuable information provided by the labels of training data. For classification tasks, supervision can help model to learn better features for classification. As for cross-domain classification tasks, previous work~\cite{bao2013partially} has shown that supervising the generative process in the source domain can help model learn better topic features for the target domain.

Following the idea of Labeled-LDA~\cite{ramage2009labeled}, our model directly assigns ground truth topic group labels for words in the source domain. While different from Labeled-LDA which performs supervision at document level (places label node $l$ at document level), we place $l$ at word level. By doing so, we can overcome the problem of Labeled-LDA that it is unable to efficiently perform sampling at test time~\cite{ramage2009labeled}. In addition, another difference of supervision used by CDL-LDA and Labeled-LDA is that Labeled-LDA performs topic level supervision while our model perform topic group level supervision.

Finally, for simplicity, we define the topic group label of a word $w$ in document $d$ of source domain as the document label of $d$.

\subsection{Inference}
The joint distribution can be decomposed as the following equation:
\begin{equation}
\begin{split}
&P(\mathbf{l},\mathbf{r},\mathbf{z},\mathbf{w}|D)\\
= &P(\mathbf{l}|D)\cdot P(\mathbf{r}|\mathbf{l},D)\cdot P(\mathbf{z}|\mathbf{r},\mathbf{l},D)\cdot P(\mathbf{w}|\mathbf{z},\mathbf{r},\mathbf{l},D) \\
=& \prod_{d}\frac{B(\eta+\mathbf{n}^l_{d})}{B(\eta)} 
\cdot \prod_{d,l}\frac{B(\gamma+\mathbf{n}^r_{l,d})}{B(\gamma)} \\
\cdot & \prod_{d,l,r}\frac{B(\alpha+\mathbf{n}^z_{r=0,l,d})}{B(\alpha)}
\cdot \prod_{d,l,r,m}\frac{B(\alpha+\mathbf{n}^z_{r=1,l,d,m})}{B(\alpha)} \\
\cdot & \prod_{l,r,z}\frac{B(\beta+\mathbf{n}^w_{z,r=0,l})}{B(\beta)}
\cdot \prod_{l,r,z,m}\frac{B(\beta+\mathbf{n}^w_{z,r=1,l,m})}{B(\beta)}
\end{split}
\end{equation}
where $B(\mathbf{v})=\frac{\prod_i\Gamma(v_i)}{\Gamma(\sum_iv_i)}$, $\mathbf{v}$ denotes vector, and $\Gamma(\cdot)$ is the gamma function. $\mathbf{n}^l_d$ is a $L$ dimension vector, and its $i$-th element is the number of times $i$-th label is seen in document $d$. Similarly, $\mathbf{n}^r_{l,d}$ is a 2 dimensional vector, and its $i$-th element is the number of times $i$-th topic type (common/specific) is seen in document $d$ and label $l$. $\mathbf{n}^z_{r=0, l, d}$ is a $T^C$ dimensional vector, and its $i$-th element is the number of times topic $i$ is assigned to common topic and label $l$ in $d$. $\mathbf{n}^z_{r=1, l, d, m}$ is a $T^S$ dimensional vector, and its $i$-th element is the number of times topic $i$ is assigned to specific topic and label $l$ in $d$, where $d$ is from domain $m$. Both $\mathbf{n}^w_{z, r=0, l}$ and $\mathbf{n}^w_{z, r=1, l, m}$ are $V$ dimensional vectors. The $i$-th element of $\mathbf{n}^w_{z, r=0, l}$ is the number of times word $i$ is assigned to common topic $z$ and label $l$. The $i$-th element of $\mathbf{n}^w_{z, r=1, l, m}$ is the number of times that word $i$ is assigned to specific topic $z$ and label $l$ in the domain $m$.

Exact inference for a complex Bayesian network is often intractable, and thus approximation methods are usually employed for inference. Blei et al \cite{blei2003latent} developed a variational EM algorithm for inference, Griffiths et al \cite{griffiths2004finding} showed how to use collapsed Gibbs sampling for approximation, which is not only simple to derive, but also can approximate to a global maximum. In this paper, we adopt collapsed Gibbs sampling to approximate the joint distribution of $l_t$, $r_t$ and $z_t$. Therefore, we have\footnote{Due to space limitation, the derivatives are ignored.}:

\begin{equation}
\centering
\begin{split}
&P(z_t=z, r_t=r, l_t=l|\mathbf{z}_{-t}, \mathbf{r}_{-t}, \mathbf{l}_{-t},\mathbf{w}, \alpha, \beta, \gamma, \eta)\\
\propto&\frac{\{N_{w_t,z,r,l}\}_{-t}+\beta}{\{N_{z,r,l}\}_{-t}+V\beta}\times\frac{\{N_{z,r,l,d}\}_{-t}+\alpha}{\{N_{r,l,d}\}_{-t}+T^r\alpha}\\
\times&\frac{\{N_{r,l,d}\}_{-t}+\gamma}{\{N_{l,d}\}_{-t}+2\cdot\gamma}\times\frac{\{N_{l,d}\}_{-t}+\eta}{\{N_d\}_{-t}+L\eta}
\end{split}
\label{inf1}
\end{equation}

\noindent where $N_x$ denotes the number of times that $x$ is observed in the corpus, and $-t$ denotes a quantity that excludes data from $t^{th}$ position. For specific meaning of each notation in the equation~\ref{inf1}, please refer to Table~\ref{tab:notations}.


\section{Experiments}
\subsection{Datasets}
\textbf{20Newsgroups}\footnote{http://qwone.com/~jason/20Newsgroups/} This dataset has been widely used for evaluating the performance of cross-domain text classification models~\cite{li2012topic,bao2013partially,pan2010cross}. It contains approximately 20,000 newsgroup documents which are organized into 20 different categories. Each category has nearly 1,000 documents. The 20 different categories can be partitioned into 7 top-categories, among which \textit{comp}, \textit{sci}, \textit{rec} and \textit{talk} have multiple sub-categories. To fairly compare with other models, we use the six cross-domain dataset (\emph{Comp vs. Rec}, \emph{Comp vs. Sci}, \emph{Comp vs. Talk}, \emph{Rec vs. Sci}, \emph{Rec vs. Talk}, \emph{Sci vs. Talk}) provided by the authors of TCA~\cite{li2012topic}. Please refer to paper of TCA~\cite{li2012topic} for more details.

\textbf{Reuters-21578}\footnote{http://www.cse.ust.hk/TL/index.html} This dataset is another popular dataset for evaluating the performance of cross-domain text classification algorithms. Dai et al~\cite{dai2007co} build three datasets (\emph{Orgs vs. People}, \emph{Orgs vs. Places}, \emph{People vs. Places}) from Reuters-21578. For more details about the dataset, please refer to their paper~\cite{dai2007co}.

\textbf{4-class 20Newsgroups} 
To test the performance of our model for multi-class cross-domain classification tasks, we generate three 4-class cross-domain datasets from the six cross-domain datasets provided by~\cite{li2012topic}. Each of the 4-class dataset is a combination of two non-overlapping datasets from the six datasets. For example, dataset \emph{Comp vs. Rec + Sci vs. Talk} is generated from dataset \emph{Comp vs. Rec} and \emph{Sci vs. Talk}. The source and the target domains of each 4-class dataset contains 8 sub-classes, respectively. We only use top-level classes as the labels of documents.  Table \ref{tab:dataset4class20News} summarizes the 4-class cross-domain datasets.

\begin{table}[t!]
\centering
\caption{4-class Datasets generated from 20Newsgroups}
\begin{tabular}{c|c|c}
\hline
Dataset & Source Domain $\mathcal{D}^s$ & Target Domain $\mathcal{D}^t$ \\
\hline
\multirow{8}{*}{\shortstack{Comp vs. Rec \\ + \\ Sci vs. Talk}} & comp.graphics& comp.os.ms-windows.misc\\
& comp.sys.ibm.pc.hardware & comp.sys.mac.hardware\\
& rec.motorcycles & rec.autos\\
& rec.sport.baseball & rec.sport.hockey\\
& sci.crypt & sci.electronics\\
& sci.med & sci.space\\
& talk.politics.misc & talk.politics.guns\\
& talk.religion.misc & talk.politics.mideast\\
\hline
\multirow{8}{*}{\shortstack{Comp vs. Sci \\ + \\ Rev vs. Talk}} & comp.os.ms-windows.misc & comp.graphics\\
& comp.sys.ibm.pc.hardware & comp.sys.mac.hardware\\
& rec.autos & rec.motorcycles\\
& rec.sport.baseball & rec.sport.hockey\\
& sci.electronics & sci.crypt\\
& sci.space & sci.med\\
& talk.politics.mideast & talk.politics.guns\\
& talk.politics.misc & talk.religion.misc\\
\hline
\multirow{8}{*}{\shortstack{Comp vs. Talk \\ + \\ Rec vs. Sci}} & comp.os.ms-windows.misc & comp.graphics\\
& comp.sys.mac.hardware & comp.sys.mac.hardware\\
& rec.autos & rec.motorcycles\\
& rec.sport.baseball & rec.sport.hockey\\
& sci.crypt & sci.electronics\\
& sci.med & sci.space\\
& talk.politics.mideast & talk.politics.guns\\
& talk.politics.misc & talk.religion.misc\\
\hline
\end{tabular}
\label{tab:dataset4class20News}
\end{table} 

\begin{table*}[t!]
\centering
\caption{Classification accuracies (\%) on 20Newsgroups and Reuters-21578}
\begin{tabular}{c|c|c|c|c|c|c|c|c|c|c}
\hline
Task & LG & SVM & SFA & TPLSA & CDPLSA & TCA & PSCCLDA & CCLDA & CDL-LDA$^{un}$ & CDL-LDA \\
\hline
Comp vs. Rec & 90.6 & 89.5 & 93.9 & 91.0 & 91.4  & 94.0 & 95.8 & 86.2 & 88.4 & \textbf{97.7}\\
Comp vs. Sci & 75.9 & 71.9 & 83.0 & 80.2 & 87.7  & 89.1 & 90.0 & 75.1 & 81.1 & \textbf{95.7}\\
Comp vs. Talk & 91.1 & 89.8 & 97.1 & 93.8  & 95.5 & 96.7 & 96.7 & 91.5 & 96.4 & \textbf{98.8}\\
Rec vs. Sci & 71.9 & 69.6 & 88.5 & 92.8  & 89.5 & 87.9 & 95.5 & 78.9 & 81.8 & \textbf{98.1}\\
Rec vs. Talk & 84.8 & 82.7 & 93.5 & 84.9  & 89.9 & 96.2 & 95.8 & 79.2 & 96.1 & \textbf{98.3}\\
Sci vs. Talk & 78.0 & 74.7 & 85.4 & 89.0 & 86.2  & 94.0 & 94.7 & 82.8 & 85.8 & \textbf{97.8}\\
\hline
Orgs vs. People & 68.1 & 67.0 & 67.1 & 74.6 & 80.8  & 79.2 & 80.7 & 66.1 & 77.4 & \textbf{84.1}\\
Orgs vs. Places & 69.2 & 66.9 & 68.3 & 71.9 & 71.4  & 73.0 & 74.2 & 54.6 & 68.0 & \textbf{76.4}\\
People vs. Places & 51.3 & 52.0 & 50.6 & 62.3 & 54.8  & 62.6 & \textbf{69.0} & 60.5 & 65.2 & 67.7\\
\hline
average & 75.7 & 73.8 & 80.8 & 82.3 & 82.9 & 85.9 & 88.0 & 75.0 & 82.2 & \textbf{90.5} \\
\hline
\end{tabular}
\label{tab:classificationAccu}
\end{table*}

\begin{table*}
\centering
\caption{Classification accuracies (\%) on 4-class datasets generated from 20Newsgroups}
\begin{tabular}{c|c|c|c|c|c|c}
\hline
Task & SVM & TCA & PSCCLDA & CCLDA & CDL-LDA$^{un}$& CDL-LDA\\
\hline
Comp vs. Rec + Sci vs. Talk & 66.5 & 67.4 & 78.7 & 56.4 & 72.8 & \textbf{85.9}\\
Comp vs. Sci + Rev vs. Talk & 63.6 & 70.7 & 82.7 & 44.6 & 66.7 & \textbf{91.8}\\
Comp vs. Talk + Rec vs. Sci & 61.3 & 75.1 & 71.6 & 51.9 & 73.9 & \textbf{89.7}\\
\hline
average & 63.8 & 71.1 & 77.7 & 51.0 & 71.1& \textbf{89.1}\\
\hline
\end{tabular}
\label{tab:4classAccu}
\end{table*}

\subsection{Baselines}
To evaluate the performances of our model on classification tasks, we compare it with two conventional classification models: Support Vector Machine (SVM) and Logistic Regression (LG); five state-of-the-art cross-domain text classification models: Spectral Feature Alignment (SFA)~\cite{pan2010cross}, Topic-bridge PLSA (TPLSA)~\cite{xue2008topic}, Collaborative Dual-PLSA (CDPLSA)~\cite{zhuang2010collaborative}, Topic Correlation Analysis (TCA)~\cite{li2012topic} and Partially Supervised Cross-Collection LDA topic model (PSCCLDA)~\cite{bao2013partially}. For binary classification tasks, the classification accuracies of these baselines reported in Table~\ref{tab:classificationAccu} are reprinted from~\cite{li2012topic,bao2013partially}. For 4-class classification tasks, the code of TCA is provided by its authors, PSCCLDA~\cite{bao2013partially} is re-implemented by ourself, and code of SVM is from LIBSVM~\cite{chang2011libsvm}.

To directly show the effectiveness of the proposed partial supervision, we also implement an unsupervised version of CDL-LDA: CDL-LDA$^{un}$. Since CDL-LDA$^{un}$ is an unsupervised model, we train a LG classifier on the source domain, and then use it to classify documents in the target domain.

In addition, to directly show the differences of proposed \emph{group alignment} and \emph{exact alignment}, we also re-implement Cross-Collection LDA (CCLDA)~\cite{paul2009cross}. The only difference between CCLDA and CDL-LDA$^{un}$ is that CCLDA adopts \emph{exact alignment} (one-to-one alignment) while CDL-LDA$^{un}$ adopts \emph{group alignment}. When performing classifications, we also adopt LG for CCLDA.

\subsection{Implementation Details} \label{implementations}
For CDL-LDA, following TCA~\cite{li2012topic}, we set the total number of topics $T^C + T^S$ to 12, 20 for the experiments on 20Newsgroups and Reuters-21578 respectively. For experiments on 4-class dataset, we set $T^C + T^S$ to be 24 since 4-class dataset is a combination of two non-overlapping datasets of 20Newsgroups. Following TCA~\cite{li2012topic} and PSCCLDA~\cite{bao2013partially}, we fix the ratio of the number of common topics as 0.5 ($T^C$ = $T^S$). The number of iterations is 50. The hyper-parameters in all experiments are set according to a grid search from dataset \emph{Comp vs. Rec}: $\alpha=10$, $\beta=0.1$, $\gamma=1$, $\eta=0.01$. 

For CDL-LDA$^{un}$ and CCLDA~\cite{paul2009cross}, we adopt the same parameter setting as CDL-LDA. As for other models in binary classification tasks, we report the their results presented in previous papers~\cite{bao2013partially,li2012topic}. For models in 4-class classification tasks, we use the same hyper-parameter settings as reported in the original papers~\cite{bao2013partially,li2012topic}. However, for comparison fairness, we also double the number of topics for these models.


\subsection{Cross-Domain Classification}
We conduct two sets of cross-domain classification experiments: binary classification (Table~\ref{tab:classificationAccu}) and 4-class classification (Table~\ref{tab:4classAccu}).  

In the binary classification tasks, we compare CDL-LDA with several state-of-the-art models: SFA~\cite{pan2010cross}, TPLSA~\cite{xue2008topic}, CDPLSA~\cite{zhuang2010collaborative}, TCA~\cite{li2012topic} and PSCCLDA~\cite{bao2013partially}. Table~\ref{tab:classificationAccu} shows that except for dataset \emph{People vs. Places}, CDL-LDA outperforms all of these state-of-the-art methods on the rest of tasks and improves the classification accuracies by $[1.7\%, 5.7\%]$. On average, CDL-LDA improves the accuracy from 88.0\% (PSCCLDA) to 90.5\%. 

In the 4-class classification tasks, we compare CDL-LDA with state-of-the-art cross-collection topic models TCA~\cite{li2012topic} and PSCCLDA~\cite{bao2013partially}. From Table \ref{tab:4classAccu}, we can observe improvements of $[7.2\%, 14.6\%]$ on different tasks, and an average improvement of $11.4\%$. These improvements indicate the effectiveness of proposed \emph{group alignment} and partial supervision. We believe the main reason is that supervision can help model to learn correct label distribution of source domain, and through \emph{group alignment} CDL-LDA can leverage the learned label distributions of the source domain to better help the learning process of the label distributions in the target domain than \emph{exact alignment}.

The following comparisons will directly show the effectiveness of the proposed \emph{group alignment} and partial supervision.

\textbf{\emph{Group alignment}.} To directly show the effectiveness of the proposed \emph{group alignment}, we implement an unsupervised version of CDL-LDA: CDL-LDA$^{un}$, and re-implement CCLDA~\cite{paul2009cross}. The only difference of CDL-LDA$^{un}$ and CCLDA is that CCLDA adopts \emph{exact alignment} while CDL-LDA$^{un}$ adopts \emph{group alignment}. As shown in Table \ref{tab:classificationAccu} and Table \ref{tab:4classAccu}, CDL-LDA$^{un}$ outperforms CCLDA on all of the tasks, and improves classification accuracies by $[2.2\%, 16.9\%]$ for binary classification tasks and $[13.1\%, 25.1\%]$ for 4-class classification tasks. Besides, CDL-LDA$^{un}$ improves averaged accuracies from 75.0\% to 82.2\% and from 51.0\% to 71.1\% on binary and 4-class classification tasks respectively.  



\textbf{Partial Supervision.} To show the effectiveness of the proposed partial supervision method, we compare CDL-LDA with CDL-LDA$^{un}$. The only difference between these two models is that CDL-LDA employs the proposed partial supervision while CDL-LDA$^{un}$ employs LG for classification after the unsupervised learning. From Table \ref{tab:classificationAccu} and Table \ref{tab:4classAccu}, we can observe significant increases of classification accuracies made by the proposed partial supervision. We can also observe improvements of $[2.2\%, 16.3\%]$ on different binary classification tasks, and improvements of $[13.1\%, 25.1\%]$ on 4-class classification tasks. On average, CDL-LDA improves classification accuracies over CDL-LDA$^{un}$ by $8.3\%$ and $18.0\%$ on on binary and 4-class classification dataset respectively.



\subsection{Perplexity}
Perplexity is a popular evaluation metric for topic models \cite{blei2003latent}, and a lower perplexity indicates a better representation or generalization ability of the model. In this paper, we adopt perplexity to evaluate models' generalization ability on the target domain. 

Perplexity is calculated through the following equation: 

\begin{equation}
\centering
P(\mathcal{D}^{tgt}|\mathcal{D}^{src}) = exp(- \frac{\sum_{d=1}^{|D^{tgt}|} \log p(\mathcal{D}_d^{tgt}|\mathcal{D}^{src})}{\sum_{d=1}^{|D^{src}|}N_d})
\label{eq:perplexity}
\end{equation}

\noindent where $\mathcal{D}^{src}$ and $\mathcal{D}^{tgt}$ are documents from the source domain and the target domain respectively; $|D^{src}|$ and $|D^{tgt}|$ are the number of documents in the source and the target domains; $N_d$ denotes the number of words in document $d$.

\begin{table}[t!]
\centering
\caption{Perplexities of different models}
\scriptsize
\begin{tabular}{@{ }c@{ }|@{ }c@{ }|@{ }c@{ }|@{ }c@{ }|@{ }c@{ }|c@{ }}
\hline
Task  & TCA & PSCCLDA & CCLDA & CDL-LDA$^{un}$ & CDL-LDA\\
\hline
Comp vs. Rec & 1054 & 1462 & 1566 & 1401 & \textbf{932}\\
Comp vs. Sci & 1302 & 1683 & 1842 & 1436 & \textbf{976}\\
Comp vs. Talk & 1269 & 1659 & 1796 & 1566 & \textbf{990}\\
Rec vs. Sci & 1329 & 1949 & 1995 & 1724 & \textbf{1126}\\
Rec vs. Talk & 1320 & 1779 & 1925 & 1690 & \textbf{1089}\\
Sci vs. Talk & 1504 & 2138 & 2199 & 1682 & \textbf{1183}\\
\hline
Orgs vs. Places & \textbf{276} & 474 & 444 & 430 & 294\\
Orgs vs. People & 298 & 427 & 393 & 386 & \textbf{263}\\
People vs. Places & 268 & 448 & 399 & 392 & \textbf{264}\\
\hline
Comp vs. Rec/Sci vs. Talk & 1077 & 1581 & 1606 & 1040 & \textbf{610}\\
Comp vs. Sci/Rev vs. Talk & 1041 & 1531 & 1564 & 1051 & \textbf{584}\\
Comp vs. Talk/Rec vs. Sci & 1055 & 1535 & 1614 & 1043 & \textbf{588}\\
\hline
average & 983 & 1389 & 1445 & 1153 & \textbf{742}\\
\hline
\end{tabular}
\label{tab:perplexity}
\end{table} 

\begin{table}[t!]
\centering
\caption{T-tests for perplexities}
\scriptsize
\begin{tabular}{@{ }l@{ }|c|c|c|c|c@{ }}
\hline
  & TCA & PSCCLDA & CCLDA & CDL-LDA$^{un}$ & CDL-LDA\\
\hline
CCL-LDA$^{un}$ & 0.0011 & 0.0010 & 0.0004 & - & $<$0.0001\\
CCL-LDA& 0.0003 & $<$0.0001 & $<$0.0001 & $<$0.0001 & -\\
\hline
\end{tabular}
\label{tab:perplexity t-test}
\end{table} 

Table~\ref{tab:perplexity} shows the perplexities of different models. To better show the improvement made by \emph{group alignment} and the proposed partial supervision, we also conduct t-test (one tail, paired)~\cite{yang1999re} (as shown in Table~\ref{tab:perplexity t-test}). The results of t-test show that both CDL-LDA and CDL-LDA$^{un}$ can consistently achieve lower perplexities than PSCCLDA and CCLDA. P-value between CDL-LDA$^{un}$ and CCLDA ($<$0.0001) indicates that \emph{group alignment} helps model to better generalize documents in the target domain than one-to-one \emph{exact alignment}. 

As for TCA, it has two steps: unsupervised feature learning \emph{without alignment}, and feature projection from the target domain into the source domain. However, perplexity scores cannot be calculated after the projection. Therefore, the perplexities of TCA presented in Table~\ref{tab:perplexity} represent the generalization ability of a cross-domain topic model which \emph{do not adopt any alignment}. In fact, we could interpret the perplexities of TCA as lower bounds of perplexities for unsupervised cross-domain topic models. This is because models without any alignment should be more flexible and can learn better representations for the target domain than the models with topic alignments. Table~\ref{tab:perplexity t-test} shows that the p-value between CDL-LDA$^{un}$ and TCA is 0.0011 ($<$0.005), which indicates that the mean perplexity of TCA is systematically lower than the mean perplexity of CDL-LDA$^{un}$. This result tells us that there is still some room for improvement. It is also interesting to find that CDL-LDA achieves lower perplexities than TCA, which indicates that the proposed partial supervision can improve the flexibility and generalization ability of the model. 

Finally, the p-value between CCL-LDA$^{un}$ and CDL-LDA is less than 0.0001, which provides another strong evidence that the proposed partial supervision can significantly help model to learn a better generalization on target domain.

\subsection{Different Numbers of Specific Topics}
To better model the different semantics of different domains and improve the model's representation flexibility, CDL-LDA allows the numbers of topics to be different for different domains. For each dataset (20Newsgroup, Reuters, generated 4-class dataset), we randomly select one task (shown in Table \ref{tab:flexibility}) to show that selecting different numbers of topics for different domains can help model to learn better representations and obtain higher classification accuracies. 

In this experiment, we fix all of the parameters as shown in section \ref{implementations}, except for the numbers of specific topics in the source and the target domains. In Table \ref{tab:flexibility}, $T^C$, $T_{src}^S$ and $T_{tgt}^S$ denote the number of common topics, the number of specific topics in the source domain, the number of specific topics in the target domain, respectively. The first rows of each task are the performances of CDL-LDA when the numbers of specific topics in both domains are the same. The second rows are the best results we can obtain by varying the number of specific topics $T_{src}^S$ and $T_{tgt}^S$.

Table \ref{tab:flexibility} shows that the accuracies are improved from 97.7\% to 98.2\%, from 67.7\% to 75.1\%, and from 85.9\% to 86.3\% on \emph{Comp vs. Rec}, \emph{People vs. Places} and \emph{Comp vs. Rec + Sci vs. Talk}, respectively. The improvements are much clear on perplexities: the perplexities are decreased by $19.0\%$, $15.5\%$ and $25.2\%$ on \emph{Comp vs. Rec}, \emph{People vs. Places} and \emph{Comp vs. Rec + Sci vs. Talk}, respectively. 

In summary, by adopting different number of topic in different domains, CDL-LDA can better model different semantics of both source and target domains, and achieve better classification performances.

\begin{table}[t!]
\centering
\caption{Case studies of different numbers of specific topics}
\scriptsize
\begin{tabular}{@{ }c@{ }|@{ }c@{ }|@{ }c@{ }|@{ }c@{ }|@{ }c@{ }|@{ }c@{ }}
\hline
Task & $T^C$ & $T_{src}^{S}$ & $T_{tgt}^{S}$ & Accuracy (\%) & Perplexity\\
\hline
\multirow{2}{*}{Comp vs. Rec} & 6 & 6 & 6 & 97.7 & 932\\
 & 6 & 8 & 3 & \textbf{98.2} & \textbf{755}\\
\hline
\multirow{2}{*}{People vs. Places} & 10 & 10 & 10 & 67.7 & 264\\
 & 10 & 13 & 7 & \textbf{75.1} & \textbf{223}\\
\hline
\multirow{2}{*}{Comp vs. Rec + Sci vs. Talk} & 12 & 12 & 12 & 85.9 & 610\\
 & 12 & 19 & 7 & \textbf{86.3} & \textbf{456}\\
\hline
\end{tabular}
\label{tab:flexibility}
\end{table} 

\subsection{Parameter Analysis}
This section presents experiments aimed at testing the influences of different parameters in CDL-LDA. We have four hyper-parameters in CDL-LDA, including $\alpha$, $\beta$, $\gamma$ and $\eta$, and two parameters about the number of topics: total number of topics $T^C+T^S$ and the ratio of common topics $T^C$. We evaluate the influence of these parameters on the 20Newsgroups dataset.

\begin{figure*}
\centering 
\subfigure[Accuracies vs. parameter $\eta$]{\label{fig:param-eta}\includegraphics[width=0.49\textwidth]{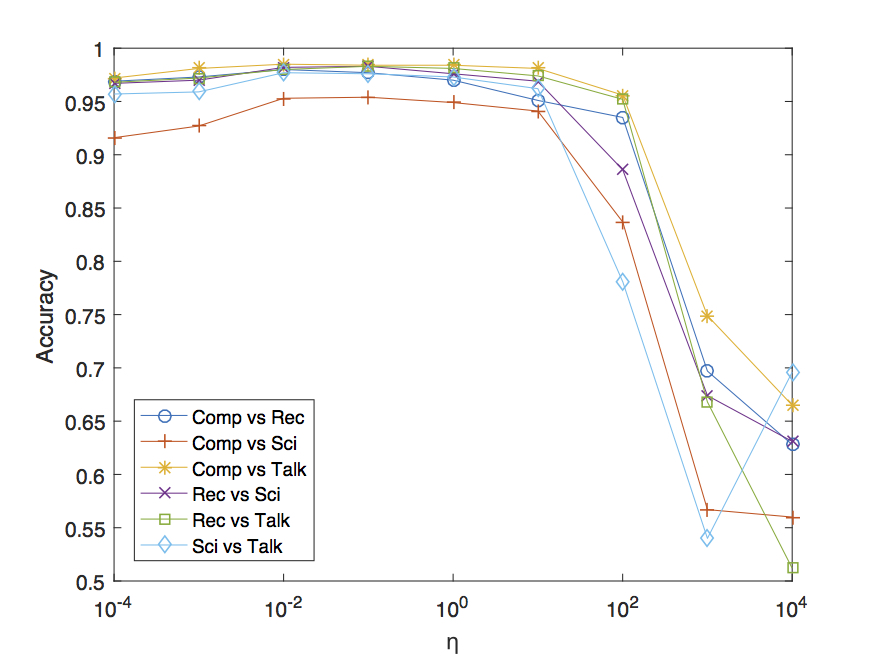}}
\subfigure[Accuracies vs. parameter $\gamma$]{\label{fig:param-gama}\includegraphics[width=0.49\textwidth]{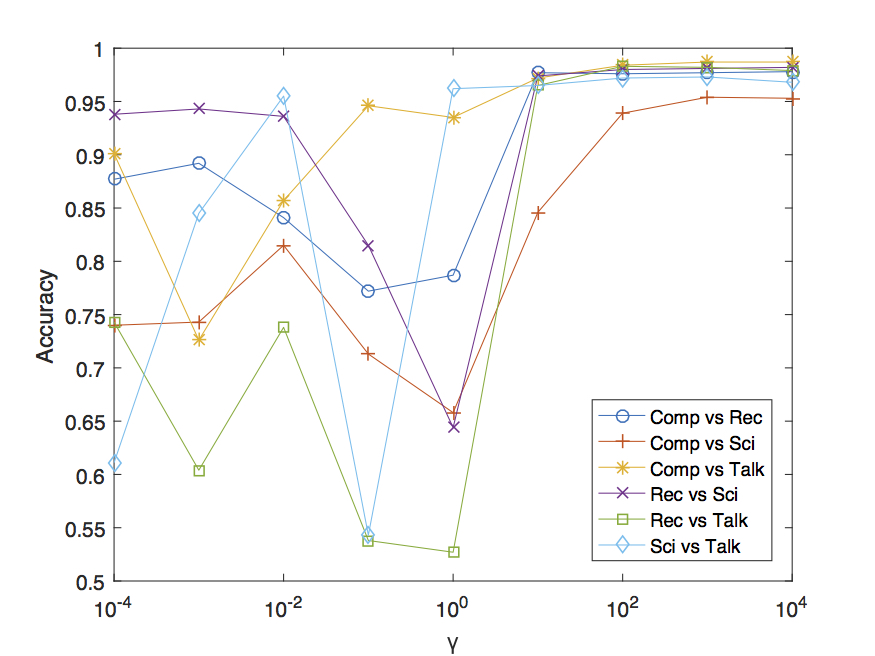}}
\subfigure[Accuracies vs. parameter $\alpha$]{\label{fig:param-alpha}\includegraphics[width=0.49\textwidth]{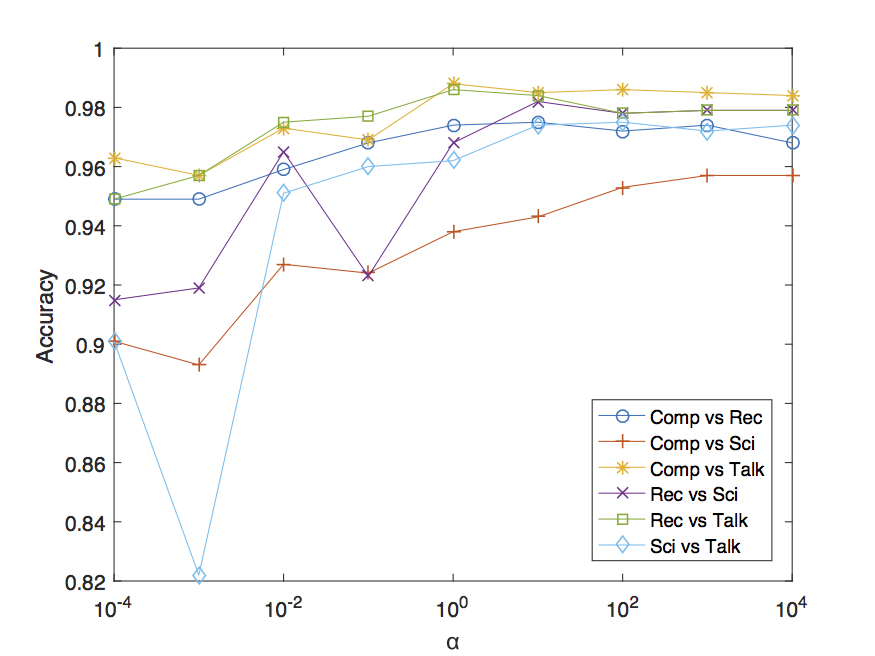}}
\subfigure[Accuracies vs. parameter $\beta$]{\label{fig:param-beta}\includegraphics[width=0.49\textwidth]{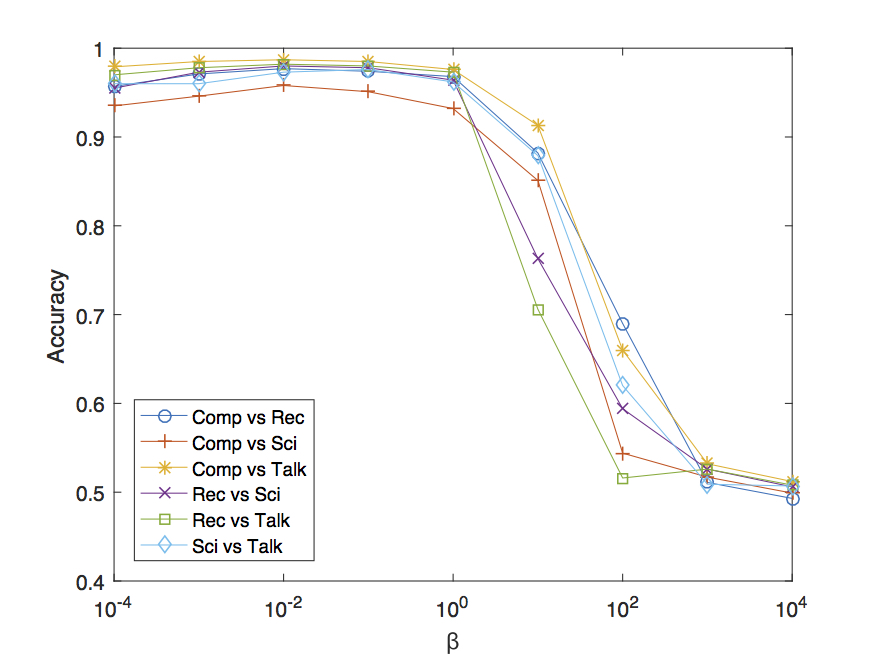}}
\subfigure[Accuracies vs. number of topics]{\label{fig:param-topics}\includegraphics[width=0.49\textwidth]{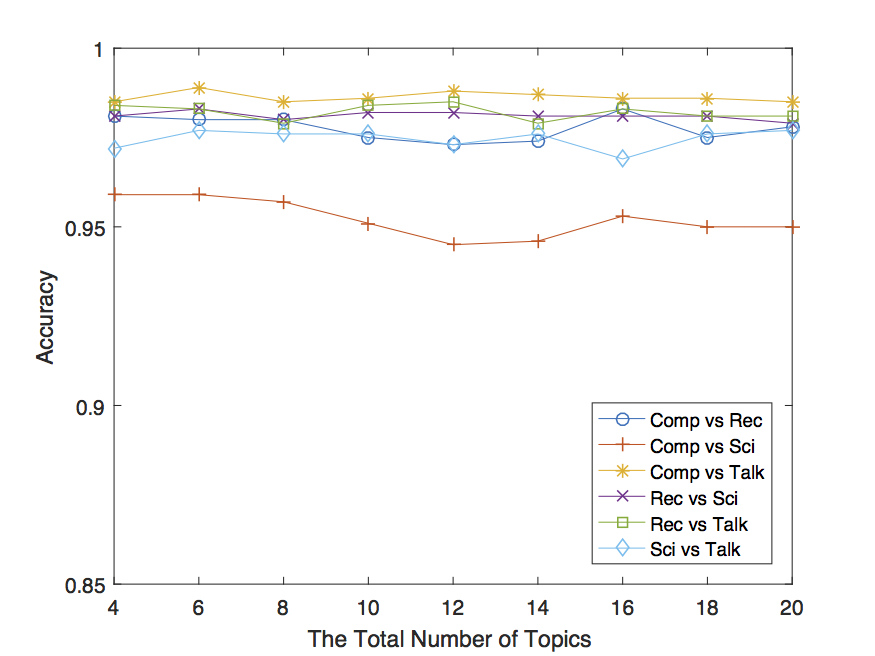}}
\subfigure[Accuracies vs. ratio of common topics]{\label{fig:param-ratios}\includegraphics[width=0.49\textwidth]{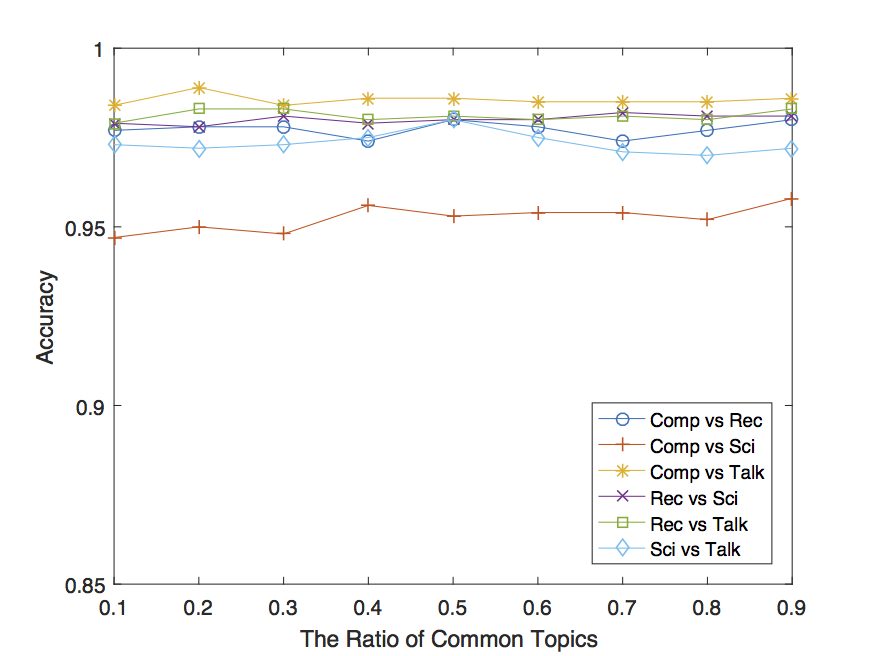}}
\caption{Parameter Analysis}
\end{figure*}

\textbf{Hyper-parameters.} 
To evaluate the individual influence of each hyper-parameter, when varying one hyper-parameter, the rest of parameters are fixed as shown in section~\ref{implementations}. 
\begin{itemize}
	\item[$\eta$:] From Fig.~\ref{fig:param-eta}, we can observe that the classification accuracies remain high and stable when $\eta\in[10^{-4}, 10]$. In theory, smaller $\eta$ encodes stronger belief that the distribution of topic groups for each document is not uniform. This result meets the fact that different documents have different labels and different topic groups. 
	\item[$\gamma$:] Fig.~\ref{fig:param-gama} shows that when $\gamma\in[10^2, 10^4]$, the classification accuracies keep high and stable, which indicates that the distribution of topic types (common/specific) is close to uniform distribution.
	\item[$\alpha$:] Fig.~\ref{fig:param-alpha} shows that a larger $\alpha$ helps model to obtain higher classification accuracies. In theory, a larger $\alpha$ implies that each document is comprised of more topics. 
	\item[$\beta$:] Fig.~\ref{fig:param-beta} shows that as $\beta\in[10^{-4}, 1]$, accuracies keep stable and high, which indicates that words in the same topic are less likely to co-occur in the same document.
\end{itemize}

\textbf{The number and the ratio of topics.}
Besides the above four hyper-parameters, the total number of topics and the ratio of common topics are two other important parameters. 

Firstly, we evaluate the influence of the total number of topics when the ratio of common topics is fixed as 0.5. As shown in Fig. \ref{fig:param-topics}, the classification accuracies of CDL-LDA are insensitive to the total number of topics. In fact, what matter most for classification accuracies are topic groups. As long as the model can correctly approximate true distributions of topic groups of documents, it can assign correct labels to these documents. 

From Fig.~\ref{fig:param-ratios}, we can observe that the ratio of the number of common topics doesn't have significant influence for classification accuracies either. Similar to previous observations, in fact, as long as the model can assign the topics of the words to correct topic groups, the model can correctly predict the labels of documents.



\begin{table*}[t!]
\centering
\caption{Topics detected by CDL-LDA on dataset \emph{Comp vs. Rec}}
\begin{tabular}{|>{\centering\arraybackslash}
p{0.75in}|>{\centering\arraybackslash}p{0.75in}|>{\centering\arraybackslash}p{0.75in}|>{\centering\arraybackslash}p{0.7in}|>{\centering\arraybackslash}p{0.7in}|>{\centering\arraybackslash}p{0.65in}|>{\centering\arraybackslash}p{0.7in}|>{\centering\arraybackslash}p{0.7in}|}
\hline
\multicolumn{4}{|c|}{Topic Group: \emph{Comp}} & \multicolumn{4}{|c|}
{Topic Group: \emph{Rec}} \\
\hline
\multicolumn{2}{|c|}{Com 1: computer science} & \multicolumn{2}{|c|}{Com 2: hardware, system} & \multicolumn{2}{|c|}{Com 3: auto} & \multicolumn{2}{|c|}{Com 4: game}\\
\hline
\multicolumn{2}{|p{1.6in}|}{edu available ftp software files program graphics system image data} 
& \multicolumn{2}{p{1.5in}|}{card don windows using drivers screen monitor able doesn buy}
& \multicolumn{2}{p{1.5in}|}{front car speed drive com ve oil miles change left} 
& \multicolumn{2}{p{1.5in}|}{game team games season play win st won teams series}\\
\hline
\centering
Src 1: graphics & Tgt 1: mac & Src 2: hardware & Tgt 2: system & Src 3: moto & Tgt 3: game& Src 4: baseball & Tgt 4: hokey\\
\hline
jpeg & apple & ide & system & dod & goal & baseball & hockey\\
image & lc & controller & mac & bike & puck & ball & nhl\\
file & power & bus & read & com & flyers& duke & gm\\
gif & centris & system & files & list & game & gant & espn\\
color & fpu & com & sys & motorcycle & leafs & ed & bruins\\
images & se & dx & network & bmw & shot & field & devils\\
format & monitor & card & re & rider & line &pitch & team\\
bit & duo & bios & disk & bikes & play & braves & playoff\\
quality & nubus & dos & file & ride & penalty & line & cup\\
version & board & board & time & motorcycles&net & east & john\\
\hline
\end{tabular}
\label{tab:topicsCCL-LDA}
\end{table*}

\begin{table*}[t!]
\centering
\caption{Classification accuracies (\%) on the dataset used by deep learning models}
\scriptsize
\begin{tabular}{c|c|c|c|c|c|c|c|c|c|c|c}
\hline
Task & LG & SVM & mSDA& $\ell_{2,1}$-SRA & DANN & ARDA& TCA & PSCCLDA & CCLDA & CDL-LDA$^{un}$ & CDL-LDA\\
\hline
Comp vs. Rec & 67.2 & 68.2   & 79.1 & 81.9 & 98.1 & 98.4  & 89.4 & 93.3 & 74.9 & 86.8 & \textbf{98.7}\\
Comp vs. Sci & 68.1 & 65.7   & 85.6 & 93.0 & 90.6 & 91.3  & 85.9 & 95.7 & 73.2 & 80.3 & \textbf{97.8}\\
Comp vs. Talk & 84.6 & 86.8  & 96.8 & 97.6 & 97.2 & 97.6  & 98.6 & 92.0 & 90.0 & 82.9 & \textbf{99.2}\\
\hline
average & 73.3 & 73.5 & 87.2 & 90.9 & 95.5 & 95.8 & 91.3 & 93.7 & 79.4 & 83.3 & \textbf{98.6}\\
\hline
\end{tabular}
\label{tab:classification3}
\end{table*}

\subsection{Topic Detection}
In this section, we qualitatively evaluate \emph{group alignment} and the proposed partial supervision adopted by CDL-LDA through topic detection experiment. In this experiment, we show four topics (indexed 1 to 4) detected by CDL-LDA in \emph{Comp vs Rec} task. In Table~\ref{tab:topicsCCL-LDA}, ``Com'', ``Src'' and ``Tgt'' refers to common topic, specific topic (source domain) and specific topic (target domain) respectively. The numbers (1 to 4) after ``Com'', ``Src'' and ``Tgt'' are topic indices.

As shown in Table \ref{tab:topicsCCL-LDA}, for topics share the same index, the concentrations of common topics, specific topics from the source and specific topics from the target domains are different. For example, for topic 3, Com 3 is about ``auto'' which includes many common words such as ``car'' and ``drive''. While Src 3 focuses on a more specific topic ``moto'' (e.g. ``bike'' and ``motorcycle''), and Tgt 3 concentrates on ``game'' (e.g. ``goal'' and ``game''). The concentration of Tgt 3 is very different from Src 3, while both of them belong to a more general topic group ``Rec''. Such behavior characterizes the \emph{group alignment}: it only align topics at topic group level instead of topic level.

\subsection{Further Comparison with Deep Learning Models}

In recent years, deep learning~\cite{lecun2015deep} has attracted a lot of attention, and many deep learning approaches have been proposed for cross-domain learning~\cite{shen2017adversarial,ganin2016domain,ICML2012Chen_416,jiang2016l2,wei2016deep,glorot2011domain,zhuang2015supervised}. In this section, we compare our model with several state-of-the-art deep learning methods on a dataset used by these deep learning models~\cite{shen2017adversarial,jiang2016l2}. The dataset is also generated from 20Newsgroup. For more details of the dataset, please refer to~\cite{shen2017adversarial,jiang2016l2}.

We select two sets of deep learning models: 1) AutoEncoder based models: marginalized Stacked Denoising AutoEncoder (mSDA)~\cite{ICML2012Chen_416} and $\ell_{2,1}$-norm Stacked Robust AutoEncoder ($\ell_{2,1}$-SRA)~\cite{jiang2016l2}; 2) Domain Adversarial Neural Network based models: Domain-Adversarial Neural Network (DANN)~\cite{ganin2016domain} and Adversarial Representation Learning for Domain Adaptation (ARDA)~\cite{shen2017adversarial}. In addition, PSCCCLDA~\cite{bao2013partially}, TCA~\cite{li2012topic}, CCLDA~\cite{paul2009cross}, LG and SVM are also adopted as baselines.

From Table~\ref{tab:classification3}, we can observe that the proposed CDL-LDA not only outperforms state-of-the-art cross-collection topic models with \emph{exact alignment}, but also the sate-of-the-art deep learning models which also adopt \emph{exact alignment}, which demonstrates the effectiveness of the proposed \emph{group alignment} and the proposed partial supervision. (Note that the classification accuracies of deep learning models in Table~\ref{tab:classification3} are reprinted from their papers~\cite{shen2017adversarial,jiang2016l2}.)

\section{Conclusion}
In this paper, we propose a novel Cross-Domain Labeled LDA (CDL-LDA) for cross-domain text classification, along with a novel \emph{group alignment} and a partial supervision. 
Different from traditional \emph{exact alignment} which directly aligns specific topics at topic level, the \emph{group alignment} aligns specific topics across domains at topic group level. Such \emph{group alignment} is guaranteed to exist and can improve model's representation flexibility.
Besides, the partial supervision directly incorporate topic group information of source domain in the training process to guide the model's learning for topic groups, which can not only reduce empirical training error on the source domain but also help the topic learning in the target domain.
Extensive quantitative experiments show that the \emph{group alignment} and the partial supervision can help model learn better features for both classification and generalization. 
Qualitative experiment shows that the proposed model is able to not only detect meaningful topics, but also successfully align topics at topic group level.

\newpage
\section*{Acknowledgment}
Dr. Deqing Wang's work was supported by the National Natural Science Foundation of China (No. 71501003).
Dr. Fuzhen Zhuang was supported by the National Natural Science Foundation of China under Grant No. 61773361, 61473273, the Project of Youth Innovation Promotion Association CAS under Grant No. 2017146. This work was also partly supported by the funding of WeChat cooperation project.

\bibliography{IEEEfull}
\bibliographystyle{IEEEtranS}

\end{document}